\documentclass[lettersize,journal]{IEEEtran}
\usepackage{amsmath,amsfonts}
\usepackage{algorithmic}
\usepackage{algorithm}
\usepackage{array}
\usepackage[caption=false,font=normalsize,labelfont=sf,textfont=sf]{subfig}
\usepackage{textcomp}
\usepackage{stfloats}
\usepackage{url}
\usepackage{verbatim}
\usepackage{graphicx}
\usepackage{cite}
\usepackage{color}
\begin{document}

\title{A Unified Object Counting Network with Object Occupation Prior}

\author{Shengqin Jiang, Qing Wang, Fengna Cheng, Yuankai Qi, Qingshan Liu,~\IEEEmembership{~IEEE Senior Member}
\thanks{Manuscript received December *** **; revised *** **, 2022. This work is supported by the National Natural Science Foundation of China (No.62001237), the Joint Funds of the National Natural Science Foundation of China (No.U21B2044), the Jiangsu Planned Projects for Postdoctoral Research Funds (No.2021K052A), the China Postdoctoral Science Foundation Funded Project (No. 2021M701756), the Startup Foundation for Introducing Talent of NUIST (No.2020r084). S. Jiang and Q. Wang contributed equally to this work. (\it{Corresponding author: Qingshan Liu, Yuankai Qi.})} 
	\thanks{S. Jiang, Q. Wang are with the School of Computer Science, Nanjing University of Information Science and Technology, Nanjing, 210044, China (e-mail: jiangshengmeng@126.com, 20211249453@nuist.edu.cn).}
	\thanks{F. Cheng is with College of Mechanical and Electronic Engineering, Nanjing Forestry University, Nanjing, 210037, China (e-mail: cfn1218@163.com)}
	\thanks{Y. Qi is with Australian Institute for Machine Learning, the University of Adelaide,
		Adelaide, SA 5005, Australia (e-mail: qykshr@gmail.com).}
	\thanks{Q. Liu are with the School of Computer Science, Nanjing University of Posts and Telecommunications, Nanjing, 210023, China (e-mail: qsliu@nuist.edu.cn).}}

\markboth{IEEE Transactions on Circuits and Systems for Video Technology XXX,~Vol.~xx, No.~x, XX~2023}%
{Shell \MakeLowercase{\textit{et al.}}: A Sample Article Using IEEEtran.cls for IEEE Journals}


\maketitle

\begin{abstract}

The counting task, which plays a fundamental role in numerous applications (e.g., crowd counting, traffic statistics), aims to predict the number of objects with various densities. Existing object counting tasks are designed for a single object class. However, it is inevitable to encounter newly coming data with new classes in our real world. We name this scenario as \textit{evolving object counting}. In this paper, we build the first evolving object counting dataset and propose a unified object counting network as the first attempt to address this task. The proposed network consists of two key components:  a class-agnostic mask module and a class-incremental module. The class-agnostic mask module learns generic object occupation prior by predicting a class-agnostic binary mask (e.g., 1 denotes there exists an object at the considering position in an image and 0 otherwise). The class-incremental module is used to handle new classes and provides discriminative class guidance for density map prediction. The combined  outputs of the class-agnostic mask module and image feature extractor are used to predict the final density map. When new classes arrive, we first add new neural nodes to the last regression and classification layers of the class-incremental module. Then, instead of retraining the model from scratch, we utilize knowledge distillation to help the model retain and consolidate what it has previously learned. We also employ a support sample bank to store a small number of typical training samples for each class, which are used to prevent the model from forgetting key information from old data. With this design, our model can efficiently and effectively adapt to new classes while maintaining good performance on already-seen data without large-scale retraining. Extensive experiments on the collected dataset demonstrate favorable performance. The dataset and code will be available at: \url{https://github.com/Tanyjiang/EOCO}.
\end{abstract}


\section{Introduction}
\IEEEPARstart{I}{n} order to analyze crowded scenarios, object counting aims to automatically estimate the number of targets in an image or a video frame~\cite{zhang2017understanding,liu2020efficient}. The importance of this task has grown as congested scenarios have become more prevalent and the need for automation has increased across industries. This task has gained widespread attention in both academia and industry, and its applications span a wide range of domains, including traffic management, smart agriculture, and public safety monitoring.



Crowd counting, as a specific application of object counting, has been extensively studied due to the prevalence of high-density scenes in public areas, gatherings, and events. In particular, the outbreak of COVID-19 in recent years has highlighted its importance and practicality~\cite{glover2004bibliometric}.  To accurately predict the number of targets, it is important to overcome various challenges, including extreme occlusion, scale variation, and cluttered background. To address these issues, Zhao et al.~\cite{zhao2019leveraging} leveraged heterogeneous auxiliary tasks, including attentive crowd segmentation, distilled depth prediction, and crowd count regression, to assist crowd counting. Zhao et al.~\cite{zhao2019scale} addressed  scale variation of pedestrians in a crowd image with the help of depth-embedded convolutional neural networks. Jiang et al.~\cite{jiang2019mask} investigated several methods for incorporating the object/non-object mask into the regression of the density map. Jiang et al.~\cite{jiang2020density} used high-level semantic information to provide effective guidance for generating high-quality density maps. Lw-Count~\cite{liu2022lw} proposed an effective counting network with a lightweight design in the encoding and decoding phases. These studies have considerably increased the performance of network prediction, which has accelerated the development of this field.


Actually, the need for counting objects in practical scenarios is common. For instance, Zhang et al.~\cite{zhang2017fcn} proposed an FCN-rLSTM network to estimate vehicle density and number; Sun et al.~\cite{sun2022wheat} detected multi-scale and dense wheat heads in a wild environment with data augmentation; and Zhou et al.~\cite{zhou2020real} developed an Android-based neural network to detect kiwifruits for yield estimation. These examples demonstrate that the class of object counting in real-world applications is dynamically changing. However, exploring class-specific counting incrementally is challenging. Training all classes together can accurately predict each class, but it requires significant hardware resources and costs when data grows exponentially. An alternative solution is fine-tuning. It facilitates the learning of new tasks by building upon previously learned knowledge. While this can lead to improved accuracy, it may also result in forgetting previously learned classes, especially as the number of classes to be learned increases. This raises the question of how to gradually learn extended classes of object counts while retaining old knowledge.


To address this challenge, we propose a unified object counting network that can handle a continuous stream of data with varying classes occurring at different time. The network comprises three modules: a base net,  a class-incremental module, and a class-agnostic mask module. This design allows the network to learn new data with new classes dynamically while also enhancing its prediction capabilities by leveraging object occupation prior. We avoid the need to retrain the network from scratch by training it with newly added data that contains new classes. Additionally, we use knowledge distillation and support sample bank to mitigate the issue of catastrophic forgetting. We carry out extensive experiments to show the validity of our suggested solutions, and we attain cutting-edge performance compared to other existing methods. The key contributions of our work are as follows:


(1) We propose a novel unified network framework for evolving object counting that, to our knowledge, is the first attempt to incrementally learn different classes of this task. To achieve this challenging task, we collected a sizable evolving object counting dataset called EoCo, which includes 6885 samples.


(2) A class-incremental module is put forward that dynamically expands the final regressor layer of the class-related density map based on the newly added class. This module enables the regressor to optimize its own parameters and select the density map while receiving guidance from the incremental classifier. 


(3) The class-agnostic mask module enhances the network's ability to perceive both interest targets and background regions by disregarding class information. This module introduces a generic object occupation prior to the network, which makes density regression less challenging. 


(4) Extensive experiments are conducted on the EoCo dataset to validate the effectiveness of our proposed strategies, and we achieve superior performance compared to recent existing methods.


\section{Related work}

\textbf{Object Counting.} Object counting aims to accurately predict the number of objects in images with varying densities. This is a valuable technology for public safety management, traffic flow monitoring, and smart agriculture~\cite{sindagi2018survey,gao2020cnn}. To better understand the progress made in this field, we can trace the evolution of the task from the perspective of crowd counting. Crowd counting models can generally be categorized into two types: traditional methods and deep learning-based methods.


Traditional methods for crowd counting include detection-based approaches~\cite{sidla2006pedestrian,subburaman2012counting} and regression-based approaches~\cite{kong2005counting,chan2008privacy}. For example, in~\cite{lin2001estimation}, a two-step approach was proposed for estimating the number of people in crowded scenes using perspective transformation. The method involved first recognizing the head-like contour and then estimating the crowd size. Meanwhile, Felzenszwalb et al.~\cite{felzenszwalb2010object} developed mixtures of multiscale deformable part models for robust object detection. Differently, Chan et al.~\cite{chan2008privacy} utilized Gaussian process regression to learn a mapping between the extracted features and the number of people per segment of the crowd. However, traditional methods rely heavily on handcrafted features that may not generalize well to different contexts.


Convolutional neural networks (CNNs) have shown excellent performance in image classification, leading to several efforts to extend their applications in crowd counting~\cite{krizhevsky2017imagenet}. One of the earliest CNN-based works in this field was \cite{zhang2015cross}, which trained a deep model to predict crowd density and count using a switchable learning procedure. However, the difficulties in network optimization arise when the crowd is too dense, making direct linear regression challenging. To address this issue, MCNN~\cite{zhang2016single} used the density map generated by a Gaussian kernel instead of directly estimating the number of individuals in the crowd. They also designed a multi-column CNN with different receptive fields to handle scale variation. Additionally, CP-CNN~\cite{sindagi2017generating} aimed to produce high-quality crowd density and count estimation by deliberately combining global and local contextual information. 

CSRNet~\cite{li2018csrnet} explored a straightforward and efficient single-column structure in contrast to the more complex multi-column network designs. It used a pre-trained VGG-16~\cite{simonyan2014very} as the backbone and stacked dilated convolutions as the backend. Zooming mechanisms for crowd counting in low- to high-density scenarios were studied by~\cite{sajid2020zoomcount}. P2PNet~\cite{song2021rethinking} exploited a purely point-based network framework for joint crowd counting and individual localization instead of predicting a density map. CCTrans~\cite{tian2021cctrans} employed a vision transformer as the backbone with multi-scale receptive fields for predicting the final results. More recently, CDANet~\cite{zhang2022cross} built a cross-domain attention network to explore the unlabeled domain on both unsupervised synthetic-to-realistic and realistic-to-realistic crowd counting. Meanwhile, FLCB~\cite{gao2022forget} studied a counting task that continuously learns from new domain data in real scenarios, rather than fitting to only one domain.

Counting various types of objects is a commonly used technique, especially in transportation and agriculture, in addition to crowd counting. For instance, Zhang et al.~\cite{zhang2017fcn} proposed a residual learning method that combines CNNs with long short-term memory  to jointly estimate the density and count of vehicles. In agriculture, Nellithimaru et al.~\cite{nellithimaru2019rols} used deep learning models and conventional 3D processing techniques to develop a pipeline for fast and accurate simultaneous localization and mapping for counting grapes.  Lins et al.~\cite{lins2020method} utilized an image processing-based method to automatically count and classify Rhopalosiphum padi. After that, Wang et al.~\cite{wang2021ssrnet} proposed a semantic segmentation regression network for counting wheat ears in remote images. These studies highlight the significance of object counting in diverse fields and the fact that target classes vary constantly in real-world applications. However, these methods usually learn a mapping for a single class, requiring the network parameters to be retrained or knowledge transfer when classes change. As a result, these methods are unable to learn the counting of dynamic object classes online, which restricts the flexibility and universality of deep-learning models.

More recently, class-agnostic counting (CAC) models have been proposed which aim to learn a unified counting model based on a few labeled exemplars. For instance, FamNet~\cite{ranjan2021learning} was the first work to achieve this task which extracts visual features from a few exemplars and matches them with those of the query image. The similarity matching outcomes are utilized as an intermediate representation to infer object counts. However, this method is severely affected by noise matching, and thus, BMNet~\cite{shi2022represent} proposed a unified similarity-aware framework that attempts to jointly learn the target representation and similarity metric. While the motivation of these methods is somewhat similar to ours, they mainly differ in three respects. Firstly, these methods rely on pre-annotated exemplars, even during the testing phase, and the accuracy of the annotations, especially box annotations, is critical for the network to reason accurately. In contrast, we do not rely on any annotation during the inference stage. Secondly, we prioritize the counting class information of the samples, whereas the CAC models tend to ignore this crucial information, which is more important in practical applications. Thirdly, the challenges faced by the two tasks are different. In other words, our model needs to learn the newly coming data of new classes without forgetting old knowledge, while the CAC models mainly focus on how to design effective similarity matching strategies.


\textbf{Class-incremental Learning.} Incremental learning refers to a system that can continuously learn new classes from fresh samples while retaining previously learned knowledge. However, a significant challenge in dynamic learning is catastrophic forgetting, which refers to the inability to recall prior knowledge while acquiring new skills. Various techniques have been developed to address this issue from different perspectives, including regularization, knowledge replay, and parameter isolation. For example, LwF~\cite{li2017learning} used a regularization strategy that imposes constraints on the loss functions of new tasks to prevent knowledge from interfering with them. In contrast,  iCaRL~\cite{rebuffi2017icarl} replayed old knowledge by reviewing a small number of representative exemplars, selected close to the class center. Additionally, parameter isolation methods~\cite{mallya2018packnet} aim to prevent knowledge forgetting by employing unique sets of parameters for each task. While most of these techniques focus on tasks such as classification~\cite{rebuffi2017icarl} and segmentation~\cite{tasar2019incremental}, they are rarely applied to regression tasks like object counting.

\section{Problem Formulation}

This section introduces the problem of class-incremental learning for object counting.  Before that, we will first go through a specific class, crowd counting, to better comprehend the main objective. There is a dataset $D = \{ {x_i},{y_i}, i=1, 2, ... , m\}$, where ${x_i}$ and ${y_i}$ denote the input image and its corresponding dot-annotation label, respectively. The label ${y_i}$ is then transformed to a density map via Gaussian kernel as the newly generated image label~\cite{zhang2016single}. To learn the mapping between inputs and labels, a network model $Net(x,\theta)$ is constructed with learnable parameters $\theta$, comprising a feature extractor $f(x,\vartheta )$ and a regressor $g(\gamma |\vartheta )$ with $\theta = \{ \vartheta ,\gamma \}$. The network parameters are optimized by minimizing the mean squared error (MSE) loss function.

\begin{equation}
	\mathcal{L} = \frac{1}{{2n}}\sum\limits_{i = 1}^n {\left\| {Net({x_i},\theta ) - {y_i}} \right\|_2^2},
\end{equation}

The aim of this study is to investigate how to effectively count inputs from a new class while still retaining previously learned information without significant degradation. In other words, the objective is to develop a well-trained network that strikes a better balance between stability and plasticity.  Without losing generality, assume that there is a dataset $D_{total}$ with ${D^{tr}}$ for training and ${D^{test}}$ for test, each of which is a member of the class set $ \mathcal{C} = \{ {c_0},{c_1},...,{c_k}\}$. The dataset is composed of $k$ counting classes, with a background class represented by ${c_0}$. This design is intended to ensure that samples that do not belong to any of the $k$ counting classes are classified as the background class.




In the given setting, the network has been trained on the sub-datasets from $D_1^{tr}$ to $D_{t-1}^{tr}$ after acquiring knowledge of the first $t-1$ classes. The earlier $t-1$ classes are referred to as old classes while learning the $t$th class. It is worth noting that $(\bigcup\limits_{i = 1}^{t - 1} {D_i^{tr})} \bigcap {D_t^{tr}} = \emptyset$, which means that there is no overlap between the training sets of the old classes and the new class. Additionally, the background samples are trained in conjunction with those of the first class. Upon completing $t$ rounds of training, the network is expected to accurately count any samples belonging to the learned classes.


\begin{figure*}[htbp]
	\centering
	\includegraphics[width=18cm]{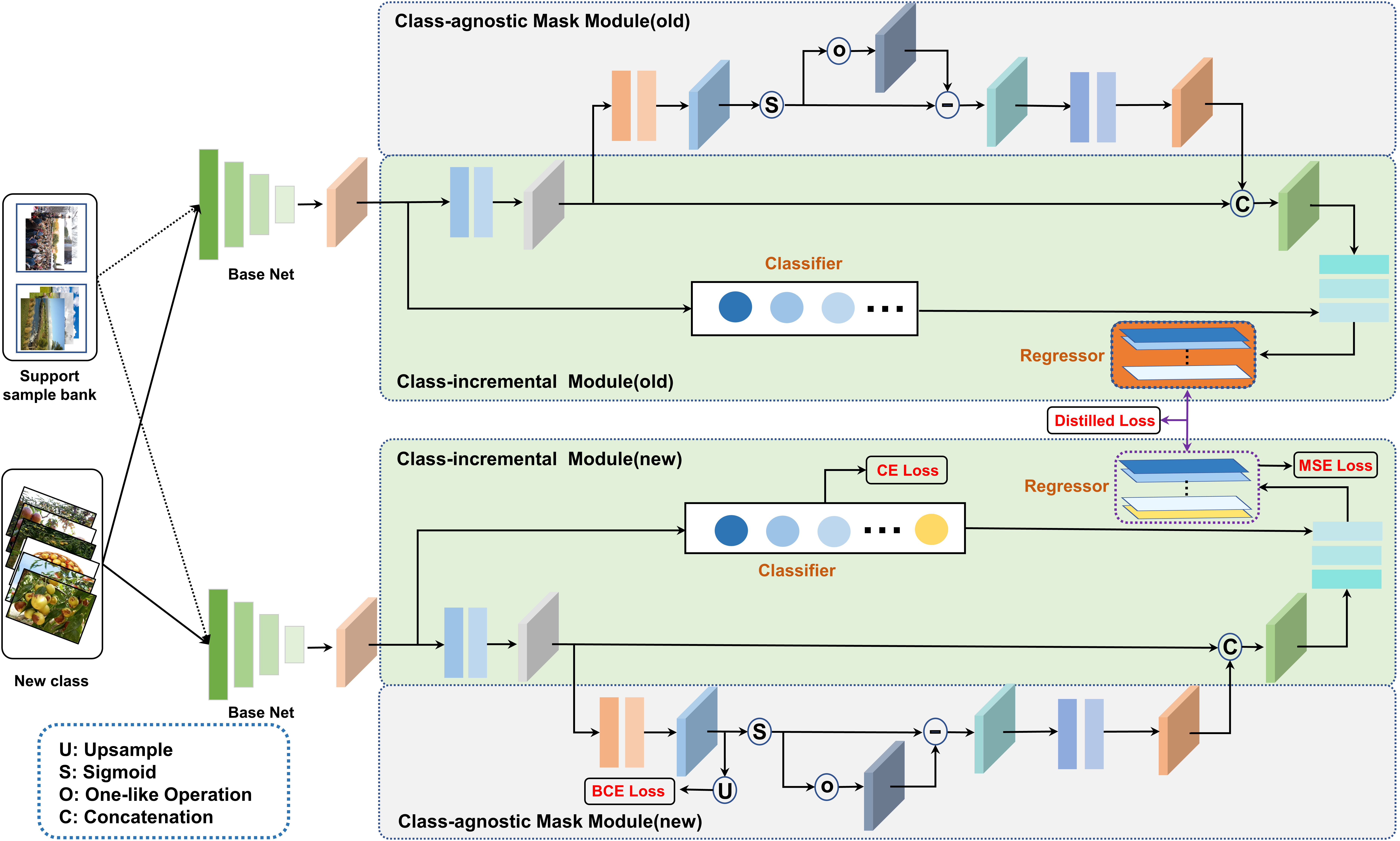}
	\caption{Overview of the $k$-th incremental step of EoCNet for object counting. The network consists of three main components: the base net, the class-agnostic mask module, and the class-incremental module. The base net is responsible for extracting discriminative features from the inputs. The class-agnostic mask module serves as an object occupation prior, enhancing the feature expression capability and allowing for better regression of the density map. The class-incremental module is designed to facilitate incremental learning of class-related regressors, which is guided online by the incremental classifier. This allows the network to adapt and learn new classes over time while alleviating forgetting previously learned ones. CE loss, BCE loss, and MSE loss denote cross entropy loss, binary cross entropy loss, and  mean squared error, respectively.}
	\label{network}
\end{figure*}

\section{Our model}

On incremental tasks such as image classification and semantic segmentation~\cite{yan2021dynamically,michieli2019incremental}, the labels all  provide explicit class information. However, object counting labels do not have such information, making it more challenging to perform class-incremental learning for this task.  As a result, adapting existing incremental learning models to address this challenge is not a straightforward process.  To overcome this difficulty, we propose a unified network framework for evolving object counting, termed EoCNet.



Fig.~\ref{network} provides an overview of the $k$-th incremental step of EoCNet for object counting. To extract features, we employ the first ten layers of VGG-16 as our network backbone. To ensure robust incremental learning at step $t$, we design three key components: a class-incremental module, a class-agnostic mask module, and a support sample bank.

\subsection{Class-incremental Module}

This module aims to dynamically extend the feature representation to be compatible with the growing number of classes. To achieve this, we use a dynamically expandable $1 \times 1$ convolutional learning representation that builds additional convolution kernels as new classes are encountered. At each step $t$, we build $t+1$ convolution kernels. The first $t$ kernel parameters are initialized using the parameters learned from the $t-1$ old classes (including the background), while the parameters of the new kernels are randomly initialized. This approach allows us to dynamically expand the prediction of density maps so that different classes have independent count outputs. If a class does not exist, it will be classified as background. However, we face a challenge that is distinct from the conventional incremental classification task. Although we explicitly sort the density map by class, there may be values in different channels of the predicted density maps if network predictions are influenced by noise. As a result, it can be difficult to determine which density map should be the output of the image.


To this end, we propose a classifier guided learning strategy  to enable the network to learn input classes and select appropriate outputs dynamically. Specifically, we begin by constructing a simple classifier. We pass the output of the feature extractor $f(x,\vartheta)$ through an average pooling layer, and then use a linear layer to make the final prediction. We optimize this classifier using the cross-entropy loss function:

\begin{equation}
	\mathcal{L}_c =  - \frac{1}{M}\sum\limits_{i = 1}^M {{y_i}\log ({p_i})},
\end{equation}\\
where $y_i $ is the ground-truth while $p_i$ refers to the corresponding predicted class label generated by the classifier, and $M$ denotes the number of training samples.


Then the learning of density map with multiple classes is defined as follows:
\begin{equation}
	{\mathcal{L}_d} = \frac{1}{{2n}}\sum\limits_{i = 1}^n {\sum\limits_{j = 1}^t {({{\hat p}_i} =  = j)} \left\| {Net({x_i},\theta )_i^j - y_i^j} \right\|_2^2},
\end{equation}
where ${{\hat p}_i} = {\text{argmax}}({p_i})$. In this manner, we are able to dynamically learn and generate density maps based on the class predictions made by the classifier. It is worth noting that, whenever a new class is encountered, the last linear layer of the class classifier is expanded to accommodate the additional class information.


\subsection{Class-agnostic Mask Module }

The class-agnostic mask module is utilized to provide a generic object occupation prior for all object classes. This module improves the extraction of regions of interest without dynamically growing the network, which reduces the complexities of network regression. Another advantage of this method is that it eliminates the need to incrementally set the convolution kernels of the module, reducing the tedium of the process and the complexity of the operations. To achieve this goal, we use a class-agnostic binary representation of the density map as supervision. Through a series of convolutional processes, we obtain a transformed feature from the output of the feature extractor $f(x,\vartheta )$. We then use a $1 \times 1$ convolution and a sigmoid operation to generate a pixel-level probability prediction. Here, we utilize the binary information of the density map as the supervision signal to learn semantic information independent of class. The loss is computed using the binary cross-entropy loss function:


\begin{equation}
	{\mathcal{L}_{a}} =  - \frac{1}{m}\sum\limits_{i = 1}^m {[{b_i}\log ({q_i}) + (1 - {b_i})\log (1 - {q_i})]},
\end{equation}\\
where ${b_i}$ and  ${q_i}$ denote the ground-truth and network prediction, respectively, $m$ denotes the total number of pixels. Note that ${q_i} = ({y_i} > \delta )$ where $\delta$ is a given threshold.

Despite applying the above supervision signal in this branch, the output  will inevitably suffer from prediction inaccuracy. To address this issue, we invert the output signal and then learn it with stacked convolution layers to reduce the impact of errors and enhance the detection of the foreground and irrelevant regions of interest. Finally, we integrate the learned features from this module with the image features to make predictions about the density map.


\subsection{Support Sample Bank}

When learning new classes, the incremental model tends to prioritize acquiring new knowledge over retaining old knowledge, leading to catastrophic forgetting. To address this issue, we build a support sample bank that leverages sample replay and knowledge distillation to recall and retain previous knowledge. Inspired by iCaRL, we follow the rehearsal principle by selecting representative exemplars from previous classes. Unlike classification-based models, we use features from intermediate convolution layers in the regressor as the sample representation. Specifically, we compress the features into a vector using an average pooling layer to represent the current sample. For each class $c_{i}$, we compute a class center by taking the average of the compressed representations of all samples. The samples are then sorted in ascending order based on their distances from their corresponding class centers. The top $K$ exemplars are chosen from the ranking list and stored as representative counting memories. It is important to note that this study uses fixed-size memory, and the number of memory samples is evenly distributed across all seen classes, including the background class.

 

As for knowledge distillation, we utilize a distillation loss to facilitate density map learning and transfer effective knowledge from the previous stage to the current stage. Specifically, the distillation loss is expressed as:

\begin{equation}
	\mathcal{L}_{kd} = \frac{1}{{2\tilde n}}\sum\limits_{i = 1}^{\tilde n} | |z_i^t - z_i^{t - 1}||_2^2,
\end{equation}
where  $z_i^{t - 1}$ and $z_i^t$ denote the predicted density estimation from the $(t-1)$th stage and the $t$th stage , respectively.

Finally, the overall objective function is given as:
\begin{equation}
	\mathcal{L}_{total}=\mathcal{L}_a + \mathcal{L}_c + (1-\lambda){\mathcal{L}_d} + \lambda \mathcal{L}_{kd},
\end{equation}
where $\lambda$ is a hyper-parameter. It is worth noting that the distillation loss function $\mathcal{L}_{kd}$ is only employed at incremental phases.

\section{Experiments}
In this section, we perform extensive experiments to illustrate the effectiveness of our proposed network. As a first attempt, we  collected an evolving counting dataset for this task, dubbed the EoCo dataset. We then conduct ablation studies to empirically showcase the impact of various factors or modules on the proposed model. Furthermore, we compare our method with the existing ones on the EoCo dataset. Before presenting the major results, we first go over the experimental settings.

\subsection{Implementation Details}

Our network is implemented using Pytorch. To prevent over-fitting during training, we resize the images to $400 \times 400$ and utilize data augmentation techniques such as random flipping, random gamma, and random grayscale. We use the Adam optimizer with a weight decay of 5e-5 to optimize the network parameters. For each incremental stage, we train the network for 300 epochs with a batch size of 8. The initial learning rate is set to 1e-5 for the base step, while the learning rate is set to 1e-5 and decreased by a factor of 10 every 100 epochs for learning new classes. The hyper-parameter $\lambda$ for the loss function is set to 0.15. To generate the ground-truth for each sample, we use a fixed-size Gaussian kernel as proposed by MCNN~\cite{zhang2016single}.


\subsection{EoCo dataset}

\begin{figure}[htbp]
	\begin{center}
		\includegraphics[width=9cm]{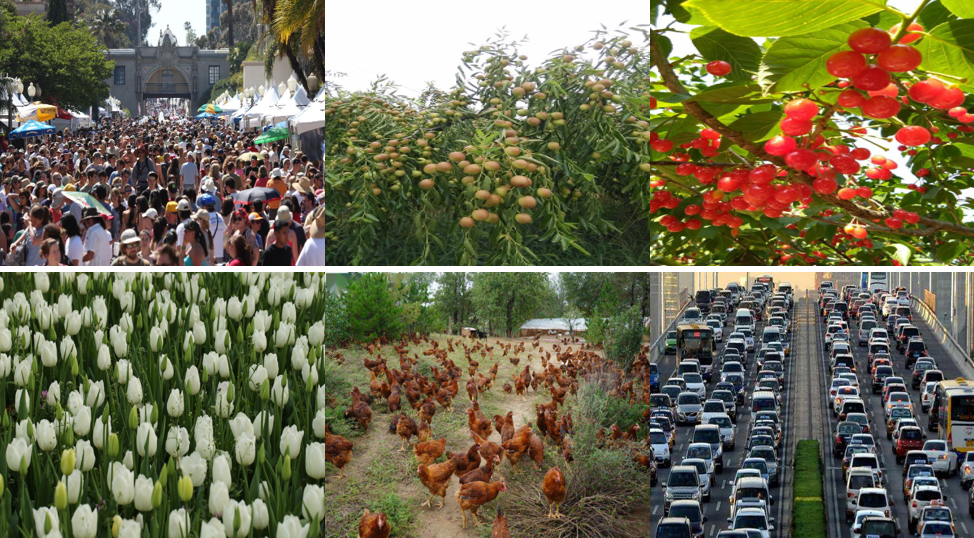}
	\end{center}
	\caption{Selected examples from EoCo Part A.}
	\label{fig2x}
\end{figure}

As there is no specialized dataset for this task, we introduce an evolving object counting dataset termed EoCo. The dataset is comprised of two parts, Part A and Part B, and includes a total of 6885 images, with 2859 images in Part A and 4026 images in Part B. The background set, which consists of 202 images, is collected from landscape images, such as grasslands and parks, and does not include any objects to be counted. Part A is divided into six classes: person, jujube, cherry, tulip, chicken, and vehicle. Some examples from this part can be seen in Fig.~\ref{fig2x}. Among these classes, the training and test of the person class are from ShanghaiTech Part A~\cite{zhang2016single}, and the remaining samples are collected using web crawlers or manual shooting. Due to the lack of a validation set in ShanghaiTech Part A, we randomly select samples from the training set for validation. Standard training, validation, and test parts are provided, ensuring that the density distribution remains as consistent as possible across all classes. A summary of the sample distribution in Part A, including the average density distribution for each class, can be found in Table~\ref{table:parta}.



On the other hand, we reorganize a dataset Part B with a larger sample size to ensure the applicability and generalizability of the considered models. All samples in the dataset are sourced from public datasets or competitions and are classified into four categories: face~\cite{yang2016wider}, wheat~\cite{kagglewheat}, person (ShanghaiTech Part B)\cite{zhang2016single}, and penguin\cite{arteta2016counting}. The dataset consists of three parts: training, validation, and test, with the sample distribution presented in Table~\ref{table:partb}. Without losing generality, we have followed the order of class-incremental learning as presented in the table.


Table~\ref{table:other_datasets} shows the comparison between our dataset and other datasets. Our dataset is notable for having a greater number of classes and a significantly larger sample size, making it well-suited for testing class-incremental tasks.

\begin{table}
	\caption[]{Sample distribution of EoCo Part A. The integer number denotes the total number of samples, and the number in brackets denotes the average density of objects in the current class.}
	\label{table:parta}
	\begin{tabular}{c|c|c|c|c}
		\hline
		Class & Train & Val & Test & Total \\
		\hline
		Person &  300 (542.36) & 59 (545.59) & 182 (433.90) & 482\\
		\hline
		Jujube & 201 (74.93) & 50 (71.12) & 248 (72.55) & 499 \\
		\hline
		Cherry & 258 (40.90) & 50 (43.24) & 182 (45.24) & 490\\
		\hline
		Tulip & 240 (65.19) & 40 (69.75) & 215 (66.25) & 495\\
		\hline
		Chicken & 220 (44.24) & 40 (47.17) & 182(46.84) & 442 \\
		\hline
		Vehicle & 231 (35.81) & 44 (34.23) & 176 (36.68) & 451 \\
		\hline
	\end{tabular}
\end{table}
\begin{table}
	\caption[h]{Sample distribution of EoCo Part B. The integer number denotes the total number of samples, and the number in brackets denotes the average density of objects in the current class.}
	\label{table:partb}
	\begin{tabular}{c|c|c|c|c}
		\hline
		Class & Train & Val & Test & Total \\
		\hline
		Face & 580 (96.45) & 60 (93.62) & 234 (98.50) & 874\\
		\hline
		Wheat & 858 (54.23) & 83 (51.87) & 372 (53.24) & 1313\\
		\hline
		Person & 400 (123.20) & 35 (129.71) & 316 (124.08) & 716 \\
		\hline
		Penguin & 740 (70.98) & 63 (79.43) & 320 (73.21) & 1123\\
		\hline
	\end{tabular}
\end{table}

\begin{table*}[t]
	\centering
	\caption[]{The comparison results of other counting datasets and EoCo dataset.}
	\label{table:other_datasets}
	\begin{tabular}[htbp]{c|c|c|c|c|c}
		\hline
		Name & Attributes  & No.classes  & No.Samples  & No.instances  & Avg.Cnt \\
		\hline
		ShanghaiTech PartA~\cite{zhang2016single}  & Free-view & 1 & 482 & 241,677 & 501  \\
		\hline
		ShanghaiTech PartB~\cite{zhang2016single} &  Surveillance-view & 1 & 716 & 88,488 & 123 \\
		\hline
		UCF-QNRF~\cite{idrees2018composition} &  Free-view & 1 & 1,535 & 1,251,642 & 815 \\
		\hline
		EoCo(Ours) &  Free-view & 9 & 6,885 & 692,980 & 101  \\
		\hline
	\end{tabular}
\end{table*}

\subsection{Evaluation Metrics}

Here we employ the mean absolute error (MAE) and the root mean squared error (MSE) as evaluation metrics. The definitions of the two metrics are presented as follows:
\begin{equation}
	MAE = \frac{1}{N}\sum\limits_{i=1}^{N}\lvert Z_i - \hat{Z}_i \rvert
\end{equation}\\
and\\
\begin{equation}
	MSE = \sqrt{\frac{1}{N}\sum\limits_{i=1}^{N}\lvert \lvert Z_i - \hat{Z}_i \rvert\rvert^2}
\end{equation}
in which $Z_i$ is the real number of the $i$th sample, $\hat{Z}_i$ is the predicted number of the $i$th sample, and $N$ is the total number of samples.

\subsection{Ablation study}

To gain more insights into our proposed method, we perform ablation studies on the important elements and components of our network on EoCo Part A.

\begin{figure}[htbp]
	\begin{center}
		\includegraphics[width=9cm]{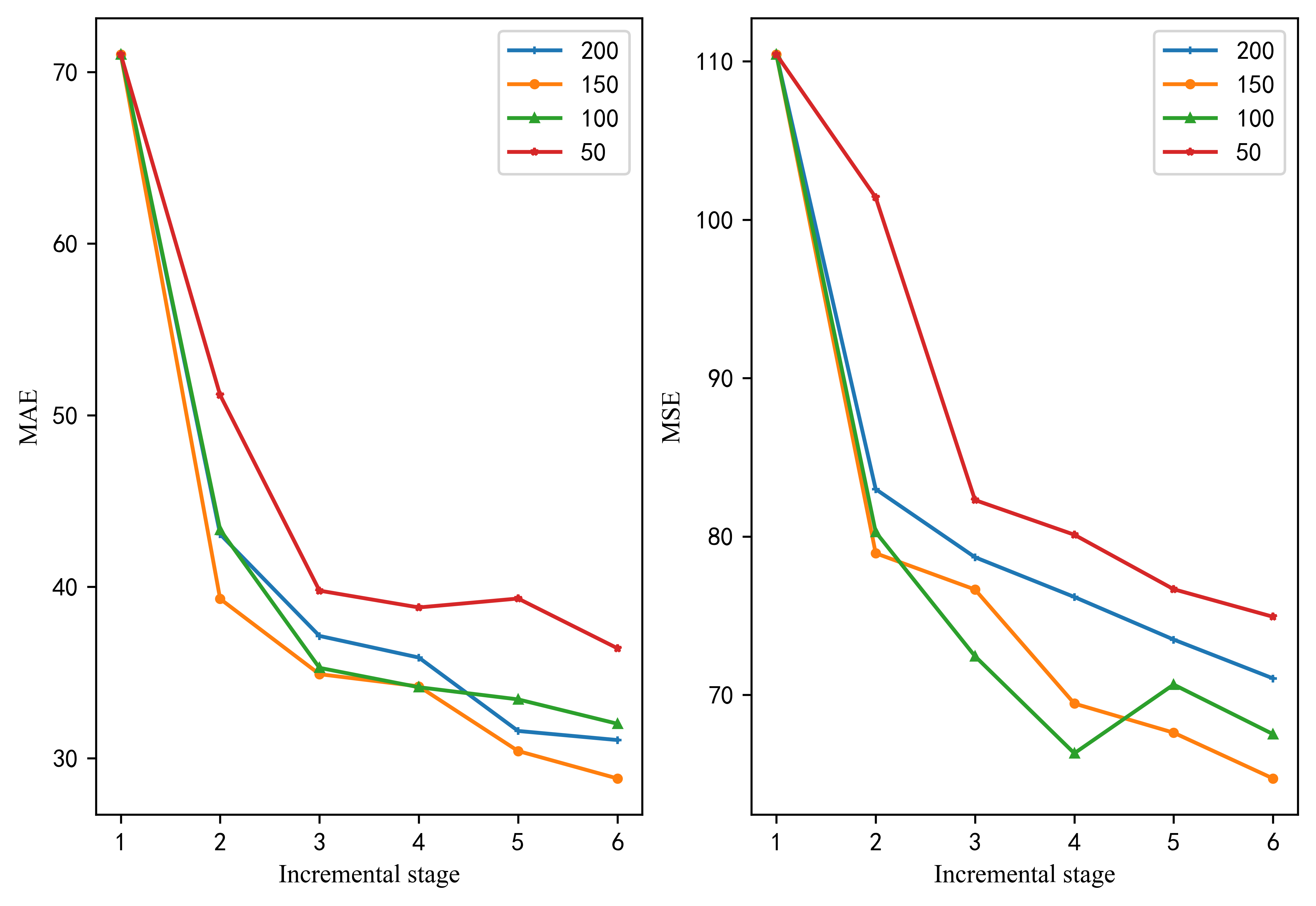}
	\end{center}
	\caption{Visualization and comparison.}
	\label{figx}
\end{figure}

(1) Effects of the quantity of memory samples

Memory samples play an important role in recovering old knowledge, of which the quantity can determine the quality of knowledge replay. Here, we study the effect of the number of memory samples on incremental learning performance, as shown in Fig.~\ref{figx}. The results indicate that, as the number of samples increases from 50 to 150, the performance (measured in terms of MAE at various incremental stages) noticeably decreases. At the numbers 100 and 150, the MSE metric shows some fluctuations from the second stage to the fifth stage, with the latter producing superior performance at the final stage. However, when the number of samples is increased to 200, the performance is not as good as that of the number 150. We speculate that this could be due to the class-centered sample selection approach not effectively considering the samples from different distributions, leading to the knowledge of other distributions being easily forgotten during subsequent learning. Consequently, we select 150 memory samples for optimal performance.


(2) Effects of backbone

We explore the effect of different network backbones on incremental models, such as MCNN~\cite{zhang2016single}, ResNet~\cite{he2016deep} and VGG16~\cite{simonyan2014very}, as shown in Tables~\ref{table:tab1} and~\ref{table:tab2}. Note that we use a three-fold wider version of MCNN to enrich the feature representation and aggregate the features from stages 2 and 3 of each ResNet model to enhance the learning of different scale features. From the tables, we can see that the performance of the network keeps getting better as the number of layers deepens, except for VGG16. This supports previous findings that deeper networks can extract features with more discriminative information~\cite{he2016deep}. Notably, the VGG16-based network achieves superior performance while utilizing 13.84 M fewer parameters than the ResNet34 and ResNet50-based networks.  This experimentally suggests that VGG16 is more adept at performing feature extraction for this task. Thus, we choose VGG-16 as the backbone of our network.
\begin{table*}[t]
	\centering
	\caption[]{Ablation study on backbone  in terms of MAE.}
	\label{table:tab1}
	\begin{tabular}{c|c|c|c|c|c|c|c}
		\hline
		Backbone & $t_0$  & $t_1$  & $t_2$  & $t_3$  & $t_4$ & $t_5$ & Param (M)\\
		\hline
		MCNN &  126.03 & 79.00 & 75.88 & 59.94 & 61.03 & 55.21  & 4.95\\
		\hline
		ResNet18 &  91.82 & 58.62 & 50.26 & 49.31 & 49.92 & 47.33 & 12.85 \\
		\hline
		ResNet34 & 85.85 & 47.01 & 43.29 & 41.25 & 50.86 & 35.86 & 18.24 \\
		\hline
		ResNet50 & 81.75 & 50.33 & 38.35 & 36.76 & 32.95 & 34.60 & 27.14 \\
		\hline
		Ours(VGG16)  & 71.03  & 39.29 & 34.90 & 34.19 & 30.42 &  28.82 & 13.84 \\
		\hline
	\end{tabular}
\end{table*}

\begin{table*}[t]
	\centering
	\caption[]{Ablation study on backbone in terms of MSE.}
	\label{table:tab2}
	\begin{tabular}{c|c|c|c|c|c|c|c}
		\hline
		Backbone & $t_0$  & $t_1$  & $t_2$  & $t_3$  & $t_4$ & $t_5$  & Param (M)\\
		\hline
		MCNN &  182.18 & 142.28 & 140.17 & 108.82 & 116.01 & 110.45 & 4.95 \\
		\hline
		ResNet18 &  144.80 & 114.36 & 93.59 & 91.05 & 92.03 & 89.73 & 12.85 \\
		\hline
		ResNet34 & 131.78 & 91.58 & 84.01 & 77.96 & 96.66 & 79.85 & 18.24 \\
		\hline
		ResNet50 & 128.33 & 107.90 & 85.50 & 77.62 & 73.16 & 70.83 & 27.14\\
		\hline
		Ours(VGG16)  & 110.44 & 78.94 & 76.65 & 69.45 & 67.61 & 64.72 & 13.84\\
		\hline
	\end{tabular}
\end{table*}

%

(3) Effects of class-agnostic mask module

To show the benefits of the class-agnostic mask module, we further conduct additional analysis of the major factors in this module and then establish the following three baselines:


\begin{itemize}
	\item \textbf{baseline 1} is a basic backbone that does not include the class-agnostic mask module. It is used to evaluate the effectiveness of this module. 
	\item \textbf{baseline 2} is the variant of the class-agnostic mask module that operates without any supervisory signals. Its purpose is to determine whether the performance enhancement is solely due to an increase in the number of network parameters.
	
	\item \textbf{baseline 3} utilizes the density map as a supervision signal instead of the binary mask. This is used to evaluate how different supervision signals affect the network's performance.
	\item \textbf{baseline 4} directly removes the convolutional learning process after mask prediction in this module. This is done to determine whether the feedback is essential for facilitating the fusion of the two features.
\end{itemize}

Table~\ref{table:tab3} presents the comparison results of our method with the different backbone architectures. Our findings are as follows:  1) Our method achieves the lowest MAE and MSE at almost all stages compared to baseline 1. This demonstrates that learning the generic object occupation prior of different classes improves the network feature representation and reduces the burden of final density regression. 2) Compared to baseline 2, the performance gain is not solely due to an increase in the number of network parameters.  3) Interestingly, baseline 3 shows that using the density map as an alternative to the binary mask also produces good results. This suggests that the two types of class-agnostic prior information can assist the network in better identifying targets, which differs from the findings in~\cite{jiang2019mask} for the crowd counting task alone. However, we note that the robustness of baseline 3 fluctuates, whereas the mask-supervised solution exhibits a monotonically decreasing trend as the number of classes increases.  4) Baseline 4 indicates that a single mask-supervised signal can impede density map regression, possibly due to the task contradiction between the two signals. The class-incremental module aims to differentiate between various classes, while this module maintains consistency in the semantics of different classes. The feedback of mask prediction bridges the gap between the two conflicting tasks through simple convolutional learning. In summary, this module can improve performance through the rational use of additional signals.

\begin{table}[t]
	\centering
	\caption[]{Ablation study on class-agnostic mask module in terms of MAE.}
	\label{table:tab3}
	\begin{tabular}{c|c|c|c|c|c|c}
		\hline
		Model & $t_0$  & $t_1$  & $t_2$  & $t_3$  & $t_4$ & $t_5$\\
		\hline
		Baseline 1  & 72.35 & 43.20 & 38.64 & 32.89 & 33.21 & 30.12 \\
		\hline
		Baseline 2 &  76.53 & 43.59 & 37.65 & 38.30 & 35.53 & 31.24 \\
		\hline
		Baseline 3 &  78.36 & 38.69 & 33.25 & 31.66 & 32.93 & 28.41 \\
		\hline
		Baseline 4 &  78.54 & 39.43 & 33.75 & 32.17 & 34.24 & 35.86 \\
		\hline
		Ours  & 71.03  & 39.29 & 34.90 & 34.19 & 30.42 &  28.82 \\
		\hline
	\end{tabular}
\end{table}

\begin{table}[t]
	\centering
	\caption[]{Ablation study on class-agnostic mask module in terms of MSE.}
	\label{table:tab4}
	\begin{tabular}{c|c|c|c|c|c|c}
		\hline
		Model & $t_0$  & $t_1$  & $t_2$  & $t_3$  & $t_4$ & $t_5$\\
		\hline
		Baseline 1  & 113.62 & 86.41 & 76.10 & 72.81 & 72.85 & 65.14 \\
		\hline
		Baseline 2 &  118.72 & 79.17 & 74.38 & 77.80 & 73.59 & 67.86 \\
		\hline
		Baseline 3 &  124.98 & 79.15 & 70.69 & 68.63 & 73.76 & 70.30 \\
		\hline
		Baseline 4 &  124.72 & 77.37 & 69.23 & 70.53 & 76.70 & 71.65\\
		\hline
		Ours  & 110.44 & 78.94 & 76.65 & 69.45 & 67.61 & 64.72 \\
		\hline
	\end{tabular}
\end{table}

%

(4)  Class-agnostic counting \it{VS} \rm{Incremental counting}

CAC aims to learn a unified model for arbitrary class counting. To compare these models with our method, we labeled exemplars with bounding boxes for each image in the test dataset. To better enable them to learn the crowded objects in our dataset, we intentionally marked boxes of three different scales, namely large, medium, and small. Tables~\ref{table:tabx1} and~\ref{table:tabx2} present a comparison of the results. It is evident that our method surpasses the two CAC methods in both metrics. The performance of FamNet~\cite{ranjan2021learning} is subpar, possibly due to the significant influence of noise on the obtained correlation maps, making it challenging to detect all the targets of each input. In comparison, BMNet+~\cite{shi2022represent} exhibits relatively good performance, demonstrating better robustness and generalization on our dataset. Apart from the advantages offered by the similarity matching method, the utilization of ResNet 50 as the backbone network plays a crucial role in attaining such performance. Here, we would like to express again that the differences between the CAC model and ours mainly lie in three aspects: (1) The CAC model generally necessitates the prior annotation of bounding boxes for object representation during reasoning, whereas our model does not require this. (2) Our model outputs the class information of the objects being counted, while the CAC model does not take this into account. (3) The challenge of our task is to learn newly coming data with new classes while not forgetting prior information, whereas the CAC model largely focuses more on how to build efficient similarity matching strategies.

\begin{table}[t]
	\centering
	\caption[]{Comparison results of the CAC methods and ours in terms of MAE on Part A.}
	\label{table:tabx1}
	\begin{tabular}{c|c|c|c|c|c|c}
			\hline
			Model & $t_0$  & $t_1$  & $t_2$  & $t_3$  & $t_4$ & $t_5$\\
			\hline
			FamNet & \multicolumn{6}{c}{64.46} \\
			\hline
			BMNet+ & \multicolumn{6}{c}{33.24} \\
			\hline
			Ours  & 71.03  & 39.29 & 34.90 & 34.19 & 30.42 &  28.82 \\
			\hline
		\end{tabular}
\end{table}

\begin{table}[t]
	\centering
	\caption[]{Comparison results of the CAC methods and ours in terms of MSE on Part A.}
	\label{table:tabx2}
	\begin{tabular}{c|c|c|c|c|c|c}
		\hline
		Model & $t_0$  & $t_1$  & $t_2$  & $t_3$  & $t_4$ & $t_5$\\
		\hline
		FamNet & \multicolumn{6}{c}{136.25} \\
		\hline
		BMNet+ & \multicolumn{6}{c}{80.66} \\
		\hline
		Ours  & 110.44 & 78.94 & 76.65 & 69.45 & 67.61 & 64.72 \\
		\hline
	\end{tabular}
\end{table}

\subsection{Evaluation on EoCo}

\begin{table}[t]
	\caption[]{Comparison results of incremental learning models in terms of MAE on Part A.}
	\centering
	\label{table:tabAA}
	\begin{tabular}{c|c|c|c|c|c|c}
		\hline
		Stage & $t_0$  & $t_1$  & $t_2$  & $t_3$ & $t_4$ & $t_5$  \\
		\hline
		FT & 78.38 & 66.04 & 78.59 & 76.60 & 67.86 & 74.03 \\
		\hline
		LwF & 86.89 & 67.73 & 75.19 &  77.70 & 73.84 & 57.22 \\
		\hline
		iCaRL & 72.35 & 43.20 & 38.64 & 32.89 & 33.21 & 30.12 \\
		\hline
		EEIL & 75.61 & 42.47 & 48.85 & 39.95 & 43.13 & 42.80 \\  
		\hline
		BiC & 68.77 & 42.07 & 39.46 & 40.36 & 45.50 & 32.67 \\
		\hline
		Ours & 71.03 & 39.29 & 34.90 & 34.19 & 30.42 & 28.82 \\
		\hline
		Joint& 71.59 & 36.97  & 30.15 & 26.20  & 24.31  & 21.58 \\
		\hline
	\end{tabular}
\end{table}

\begin{table}[t]
	\caption[]{Comparison results of incremental learning models in terms of MSE on Part A.}
	\centering
	\label{table:tabAS}
	\begin{tabular}{c|c|c|c|c|c|c}
		\hline
		Stage & $t_0$  & $t_1$  & $t_2$  & $t_3$ & $t_4$ & $t_5$  \\
		\hline
		FT & 121.59 & 133.95 & 165.01 & 169.63 & 151.14 & 166.15 \\
		\hline
		LwF & 137.51 & 136.87 & 156.87 &  175.73 & 149.75 & 128.57 \\
		\hline
		iCaRL & 113.62 & 86.41 & 76.10 & 72.81 & 72.85 & 65.14 \\
		\hline
		EEIL & 117.55 & 77.95 & 97.27 & 82.28 & 86.90 & 83.77 \\  
		\hline
		BiC & 111.05 & 76.26 & 79.63 & 78.43 & 90.28 & 77.72 \\
		\hline
		Ours & 110.44 & 78.94 & 76.65 & 69.45 & 67.61 & 64.72 \\
		\hline
		Joint& 114.24 & 70.68  & 30.15 & 56.30  & 55.68  & 51.66 \\
		\hline
	\end{tabular}
\end{table}

In this subsection, we will demonstrate the superiority of our method by comparing it with some incremental learning methods. However, since most incremental works are focused on classification tasks, we will reproduce their methods using the framework of counting tasks to enable a fair comparison. The following methods will be used for comparison:

(1) iCaRL~\cite{rebuffi2017icarl}. As mentioned previously, iCaRL serves as our baseline 1, which includes an incremental classifier as well as knowledge distillation and prototype rehearsal for representational learning. In addition, our setup includes an incremental regression layer for density map prediction.


(2) LwF~\cite{li2017learning}. In contrast to iCaRL, LwF does not have any memory available. Instead, the new task classifiers and regression layers are trained using examples from new tasks during the training phase, and all classifiers and regression layers are subsequently fine-tuned using the same instances.


(3) Fine-tuning (FT). Following the operation in~\cite{belouadah2019il2m}, we fine-tune the whole network after adding a new FC layer and a new regression output.

(4) EEIL~\cite{castro2018end}. This work proposed a combination of a cross-distilled loss and a representative memory component to maintain knowledge from old classes. Furthermore, a balanced fine-tuning strategy was introduced to address the issue of unbalanced training conditions.


(5) BiC~\cite{wu2019large}. In accordance with the setup outlined in this work, both the examples from the old class and the samples from the new class are divided into training and validation sets. The validation set is then utilized for bias correction. Due to the limited number of samples in the exemplars, the training/validation split ratio is set at 2:1.


(6) Joint. Samples from different classes are jointly optimized to train the network parameters, which can be considered as the upper limit of incremental learning performance.


\begin{figure*}[htbp]
	\centering
	\includegraphics[width=17.9cm]{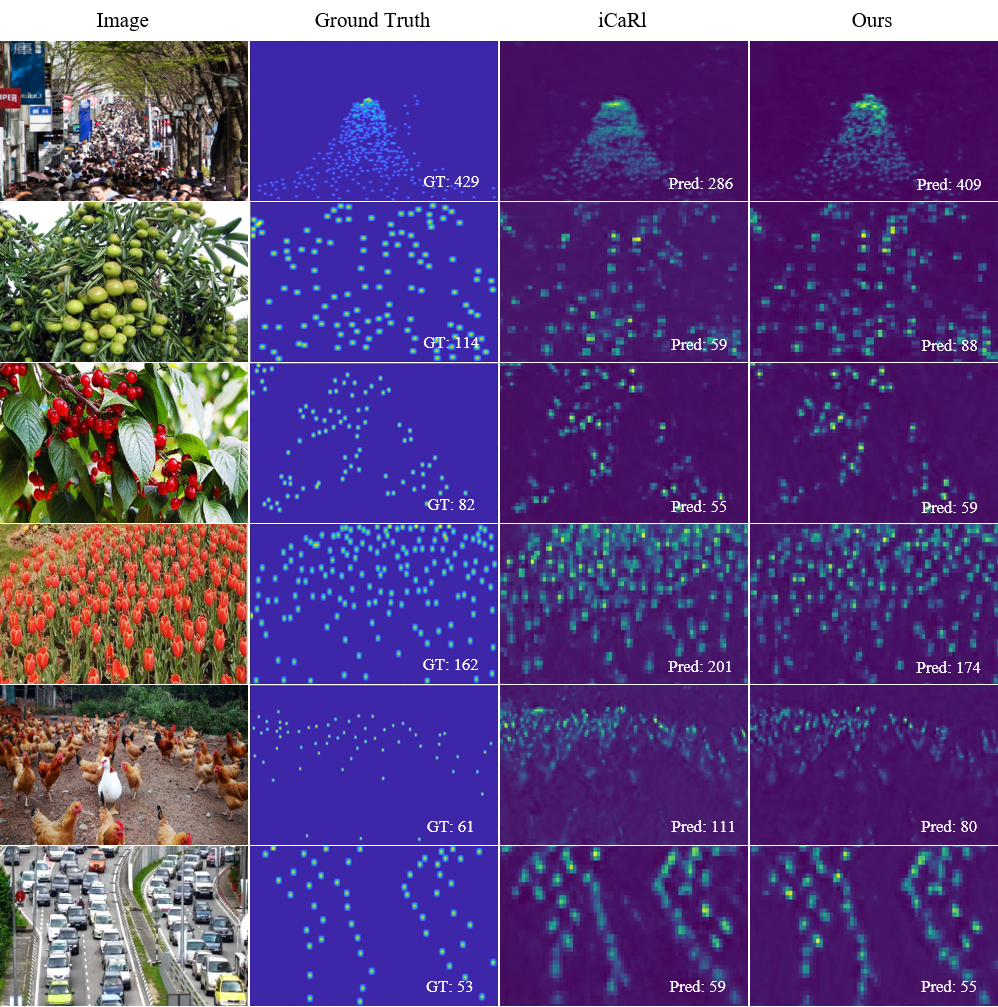}
	\caption{Visualization results. The first column shows the raw images from the final incremental stage in Part A. The second column shows the ground truth of the images in first column. The last two columns are the estimated density map by iCaRL and our method. The oldest class is at the top, followed by the new class in order.}
	\label{vis}
\end{figure*}

We perform comparative experiments on the EoCo dataset using the aforementioned methods. To ensure a fair comparison, we set the exemplar number to be 150 for rehearsal-based methods. The comparison results for Part A are summarized in Tables~\ref{table:tabAA} and \ref{table:tabAS}. As indicated, our proposed method outperforms the existing incremental methods on two evaluation metrics at nearly every stage. In contrast, LwF and FT do not perform well, primarily due to their inability to effectively retain old knowledge while focusing on learning new knowledge. On the other hand, rehearsal-based methods, such as BiC and iCaRL, are better equipped to resist catastrophic forgetting by consistently reviewing examples from previous classes. Although BiC has achieved promising results among the rehearsal-based methods, it is a two-stage method that requires more training time. 


In Part B, the number of samples in each class has significantly increased, making it much more challenging for networks to retain old knowledge using the same exemplars from Part A. As shown in Tables~\ref{table:tabBA} and \ref{table:tabBS}, our method consistently outperforms the other methods in all incremental training phases. Similar to Part A, FT and LwF perform poorly in comparison to rehearsal-based methods. Notably, our method outperforms iCaRL, particularly in the final stage. This indicates that our method is capable of extracting more discriminative features and achieving a better balance between new knowledge learning and old knowledge retention. In summary, our method is highly effective for the incremental learning task of crowd counting.


\begin{table}[htb]
	\centering
	\caption[]{Comparison results of incremental learning models in terms of MAE on Part B.}
	\centering
	\label{table:tabBA}
	\begin{tabular}{c|c|c|c|c}
		\hline
		Stage & $t_0$  & $t_1$  & $t_2$  & $t_3$  \\
		\hline
		FT & 23.24 & 29.99 & 25.58 & 37.58 \\
		\hline
		LwF & 28.21  & 29.15 & 29.75 & 44.18 \\
		\hline
		iCaRL & 23.62 & 13.65 & 12.90 & 19.24 \\
		\hline
		EEIL & 24.57 & 13.76 & 17.37 & 24.98 \\
		\hline
		BiC & 24.13 & 14.66 & 23.49 & 27.04 \\
		\hline
		Ours &  23.73 & 13.15 & 12.25 & 16.97 \\
		\hline
		Joint & 24.08 & 12.38 & 11.65 & 12.83 \\
		\hline
	\end{tabular}
\end{table}

\begin{table}[htb]
	\caption[]{Comparison results of incremental learning models in terms of MSE on Part B.}
	\centering
	\label{table:tabBS}
	\begin{tabular}{c|c|c|c|c}
		\hline
		Stage & $t_0$  & $t_1$  & $t_2$  & $t_3$  \\
		\hline
		FT & 55.59 & 79.01 & 53.16 & 73.17 \\
		\hline
		LwF & 64.25  & 64.95 & 54.44 & 82.65 \\
		\hline
		iCaRL & 55.86 & 41.09 & 30.01 & 37.37 \\
		\hline
		EEIL & 57.85 & 38.63 & 33.76 & 41.31 \\
		\hline
		BiC & 55.75 & 36.87 & 41.94 & 47.84 \\
		\hline
		Ours &  57.37 & 40.27 & 28.91 & 31.67 \\
		\hline
		Joint & 56.78 & 35.80 & 29.76 & 28.55 \\
		\hline
	\end{tabular}
\end{table}

%

\begin{table*}[t]
	\centering
	\caption[]{Comparison results of incremental learning models in terms of MAE on the entire dataset.}
	\label{table:all_mae}
	\begin{tabular}{c|c|c|c|c|c|c|c|c|c}
		\hline
		Stage & $t_0$  & $t_1$  & $t_2$  & $t_3$  & $t_4$ & $t_5$ & $t_6$  & $t_7$  & $t_8$ \\
		\hline
		iCaRL  & 23.62 & 13.65 & 12.90 & 19.24 & 22.59 & 21.44 & 25.83 & 22.14 & 24.48  \\
		\hline
		Ours &  23.73 & 13.15 & 12.25 & 16.97 & 22.18 & 20.00 & 22.53 & 20.52 & 22.63 \\
		\hline
	\end{tabular}
\end{table*}

\begin{table*}[t]
	\centering
	\caption[]{Comparison results of incremental learning models in terms of MSE on the entire dataset.}
	\label{table:all_mse}
	\begin{tabular}{c|c|c|c|c|c|c|c|c|c}
		\hline
		Stage & $t_0$  & $t_1$  & $t_2$  & $t_3$  & $t_4$ & $t_5$ & $t_6$  & $t_7$  & $t_8$ \\
		\hline
		iCaRL  & 55.86 & 41.09 & 30.01 & 37.37 & 40.43 & 39.72 & 46.22 & 40.61 & 44.10  \\
		\hline
		Ours &  57.37 & 40.27 & 28.91 & 31.67 & 38.62 & 36.00 & 40.92 & 38.15 & 41.23 \\
		\hline
	\end{tabular}
\end{table*}

\begin{table}[t]
	\centering
	\caption[]{Comparison results of the domain-incremental tasks on MAE (MSE).}
	\label{table:domaintask}
	\begin{tabular}{c|c|c|c}
		\hline
		Stage & $t_0$  & $t_1$  & $t_2$  \\
		\hline
		iCaRL & 68.32 (107.18) & 92.81 (160.81) & 94.84 (189.20)  \\
		\hline
		Ours &  66.88 (107.45) & 90.02 (154.01) & 70.52 (143.38) \\
		\hline
	\end{tabular}
\end{table}

To further explore the incremental performance of our model on additional classes, we have employed the model trained on Part B to continue incremental class learning on Part A. Here, we exclude the person class from this new incremental learning process since it is already included in part B. Our results, which are presented in Tables~\ref{table:all_mae} and~\ref{table:all_mse}, consistently show that our method outperforms iCaRL as we continue to learn new classes. This indicates that our proposed method has a robust ability to facilitate continuous learning and resist forgetting.

In addition, we study the performance of our proposed method on a domain-incremental task of crowd counting by selecting three commonly used counting datasets (ShanghaiTech Part A, UCF-QNRF, and ShanghaiTech Part B) and learning them in a sequential order. As our proposed method is not specifically designed for a domain-incremental task, we remove the class-incremental module from our network and employ a single regression head. As shown in Table~\ref{table:domaintask}, our method outperforms iCaRL, demonstrating the effectiveness of our method for this visual task.

\subsection{Visualization}

To provide additional evidence for the effectiveness of our method, we compare the estimated density maps of our approach to iCaRL at the last incremental stage, as illustrated in Fig.~\ref{vis}. The results demonstrate that our method produces a more accurate count estimation than iCaRL. For instance, in the first image, our method provides a more precise count estimate for extremely dense crowds compared to iCaRL, which may struggle to retain old information. This observation is also evident in other classes. Overall, these results suggest that our approach exhibits higher memory retention and learning capacity.



\begin{figure}[htbp]
	\centering
	\includegraphics[width=9cm]{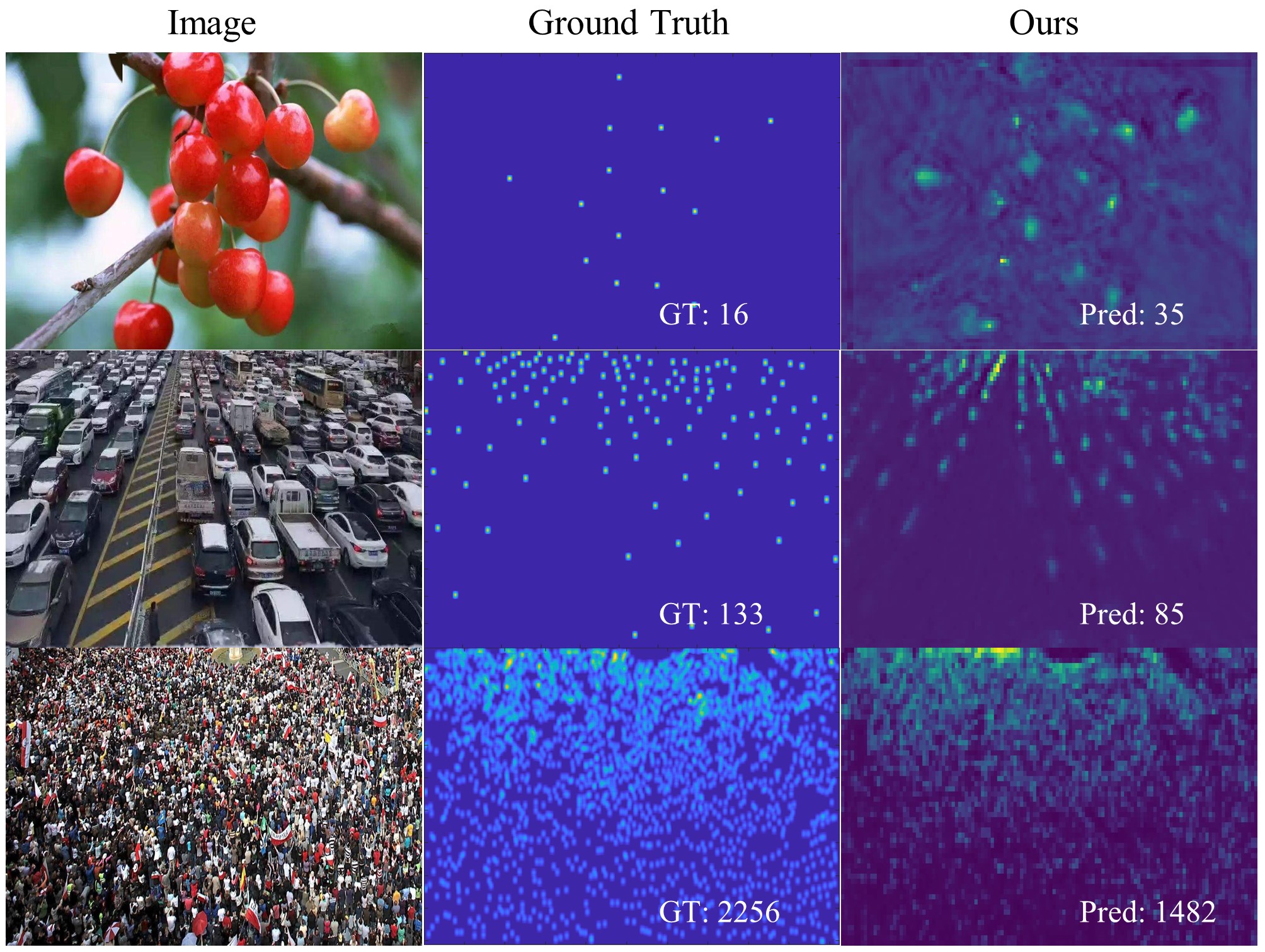}
	\caption{Negative results of our method at the final incremental stage.}
	\label{vis_f}
\end{figure}

Additionally, we present some negative results and identify potential factors contributing to these outcomes, as demonstrated in Fig.~\ref{vis_f}. The figure reveals that the network tends to overestimate counts at low densities, while underestimating them at higher densities. We attribute this issue to two primary factors. On the one hand, catastrophic forgetting during the incremental process can degrade the prediction performance of existing categories. On the other hand, the limited feature extraction capability of the backbone network may make the model less resilient to challenges such as severe occlusions and complex backgrounds.


\section{Conclusion}

In this study, we introduce a novel unified evolving object counting network for counting evolving objects. The network comprises three components: the base net, the class-agnostic mask module, and the class-incremental module. The base net extracts discriminative features from the inputs, while the class-agnostic mask module serves as an object occupation prior to enhancing the feature expression capability, thereby enabling better regression of the density map. The class-incremental module implements incremental learning of class-related regressors, guided online by the incremental classifier. We also employ the knowledge distillation and support sample bank to better retain old knowledge at feature and image levels, respectively. To evaluate our method, we collected a new class-incremental dataset called EoCo, covering a range of density scenarios.  Extensive experiments demonstrate the efficacy of our proposed method.




\bibliographystyle{IEEEtran}
\bibliography{reference}

\begin{thebibliography}{10}
\providecommand{\url}[1]{#1}
\csname url@samestyle\endcsname
\providecommand{\newblock}{\relax}
\providecommand{\bibinfo}[2]{#2}
\providecommand{\BIBentrySTDinterwordspacing}{\spaceskip=0pt\relax}
\providecommand{\BIBentryALTinterwordstretchfactor}{4}
\providecommand{\BIBentryALTinterwordspacing}{\spaceskip=\fontdimen2\font plus
\BIBentryALTinterwordstretchfactor\fontdimen3\font minus
  \fontdimen4\font\relax}
\providecommand{\BIBforeignlanguage}[2]{{%
\expandafter\ifx\csname l@#1\endcsname\relax
\typeout{** WARNING: IEEEtran.bst: No hyphenation pattern has been}%
\typeout{** loaded for the language `#1'. Using the pattern for}%
\typeout{** the default language instead.}%
\else
\language=\csname l@#1\endcsname
\fi
#2}}
\providecommand{\BIBdecl}{\relax}
\BIBdecl

\bibitem{zhang2017understanding}
S.~Zhang, G.~Wu, J.~P. Costeira, and J.~M. Moura, ``Understanding traffic
  density from large-scale web camera data,'' in \emph{Proceedings of the IEEE
  Conference on Computer Vision and Pattern Recognition}, 2017, pp. 5898--5907.

\bibitem{liu2020efficient}
L.~Liu, J.~Chen, H.~Wu, T.~Chen, G.~Li, and L.~Lin, ``Efficient crowd counting
  via structured knowledge transfer,'' in \emph{Proceedings of the ACM
  International Conference on Multimedia}, 2020, pp. 2645--2654.

\bibitem{glover2004bibliometric}
S.~W. Glover and S.~L. Bowen, ``Bibliometric analysis of research published in
  tropical medicine and international health 1996--2003,'' \emph{Tropical
  Medicine \& International Health}, vol.~9, no.~12, pp. 1327--1330, 2004.

\bibitem{zhao2019leveraging}
M.~Zhao, J.~Zhang, C.~Zhang, and W.~Zhang, ``Leveraging heterogeneous auxiliary
  tasks to assist crowd counting,'' in \emph{Proceedings of the IEEE/CVF
  Conference on Computer Vision and Pattern Recognition}, 2019, pp.
  12\,736--12\,745.

\bibitem{zhao2019scale}
M.~Zhao, C.~Zhang, J.~Zhang, F.~Porikli, B.~Ni, and W.~Zhang, ``Scale-aware
  crowd counting via depth-embedded convolutional neural networks,'' \emph{IEEE
  Transactions on Circuits and Systems for Video Technology}, vol.~30, no.~10,
  pp. 3651--3662, 2019.

\bibitem{jiang2019mask}
S.~Jiang, X.~Lu, Y.~Lei, and L.~Liu, ``Mask-aware networks for crowd
  counting,'' \emph{IEEE Transactions on Circuits and Systems for Video
  Technology}, vol.~30, no.~9, pp. 3119--3129, 2019.

\bibitem{jiang2020density}
X.~Jiang, L.~Zhang, T.~Zhang, P.~Lv, B.~Zhou, Y.~Pang, M.~Xu, and C.~Xu,
  ``Density-aware multi-task learning for crowd counting,'' \emph{IEEE
  Transactions on Multimedia}, vol.~23, pp. 443--453, 2020.

\bibitem{liu2022lw}
Y.~Liu, G.~Cao, H.~Shi, and Y.~Hu, ``Lw-count: An effective lightweight
  encoding-decoding crowd counting network,'' \emph{IEEE Transactions on
  Circuits and Systems for Video Technology}, 2022.

\bibitem{zhang2017fcn}
S.~Zhang, G.~Wu, J.~P. Costeira, and J.~M. Moura, ``Fcn-rlstm: Deep
  spatio-temporal neural networks for vehicle counting in city cameras,'' in
  \emph{Proceedings of the IEEE International Conference on Computer Vision},
  2017, pp. 3667--3676.

\bibitem{sun2022wheat}
J.~Sun, K.~Yang, C.~Chen, J.~Shen, Y.~Yang, X.~Wu, and T.~Norton, ``Wheat head
  counting in the wild by an augmented feature pyramid networks-based
  convolutional neural network,'' \emph{Computers and Electronics in
  Agriculture}, vol. 193, p. 106705, 2022.

\bibitem{zhou2020real}
Z.~Zhou, Z.~Song, L.~Fu, F.~Gao, R.~Li, and Y.~Cui, ``Real-time kiwifruit
  detection in orchard using deep learning on android™ smartphones for yield
  estimation,'' \emph{Computers and Electronics in Agriculture}, vol. 179, p.
  105856, 2020.

\bibitem{sindagi2018survey}
V.~A. Sindagi and V.~M. Patel, ``A survey of recent advances in cnn-based
  single image crowd counting and density estimation,'' \emph{Pattern
  Recognition Letters}, vol. 107, pp. 3--16, 2018.

\bibitem{gao2020cnn}
G.~Gao, J.~Gao, Q.~Liu, Q.~Wang, and Y.~Wang, ``Cnn-based density estimation
  and crowd counting: A survey,'' \emph{arXiv preprint arXiv:2003.12783}, 2020.

\bibitem{sidla2006pedestrian}
O.~Sidla, Y.~Lypetskyy, N.~Brandle, and S.~Seer, ``Pedestrian detection and
  tracking for counting applications in crowded situations,'' in
  \emph{Proceedings of the IEEE International Conference on Video and Signal
  Based Surveillance}.\hskip 1em plus 0.5em minus 0.4em\relax IEEE, 2006, pp.
  70--70.

\bibitem{subburaman2012counting}
V.~B. Subburaman, A.~Descamps, and C.~Carincotte, ``Counting people in the
  crowd using a generic head detector,'' in \emph{Proceedings of the
  International Conference on Advanced Video and Signal-based
  Surveillance}.\hskip 1em plus 0.5em minus 0.4em\relax IEEE, 2012, pp.
  470--475.

\bibitem{kong2005counting}
D.~Kong, D.~Gray, and H.~Tao, ``Counting pedestrians in crowds using viewpoint
  invariant training.'' in \emph{Proceedings of the British Machine Vision
  Conference}, vol.~1.\hskip 1em plus 0.5em minus 0.4em\relax Citeseer, 2005,
  p.~2.

\bibitem{chan2008privacy}
A.~B. Chan, Z.-S.~J. Liang, and N.~Vasconcelos, ``Privacy preserving crowd
  monitoring: Counting people without people models or tracking,'' in
  \emph{Proceedings of the IEEE Conference on Computer Vision and Pattern
  Recognition}.\hskip 1em plus 0.5em minus 0.4em\relax IEEE, 2008, pp. 1--7.

\bibitem{lin2001estimation}
S.-F. Lin, J.-Y. Chen, and H.-X. Chao, ``Estimation of number of people in
  crowded scenes using perspective transformation,'' \emph{IEEE Transactions on
  Systems, Man, and Cybernetics-Part A: Systems and Humans}, vol.~31, no.~6,
  pp. 645--654, 2001.

\bibitem{felzenszwalb2010object}
P.~F. Felzenszwalb, R.~B. Girshick, D.~McAllester, and D.~Ramanan, ``Object
  detection with discriminatively trained part-based models,'' \emph{IEEE
  Transactions on Pattern Analysis and Machine Intelligence}, vol.~32, no.~9,
  pp. 1627--1645, 2010.

\bibitem{krizhevsky2017imagenet}
A.~Krizhevsky, I.~Sutskever, and G.~E. Hinton, ``Imagenet classification with
  deep convolutional neural networks,'' \emph{Communications of the ACM},
  vol.~60, no.~6, pp. 84--90, 2017.

\bibitem{zhang2015cross}
C.~Zhang, H.~Li, X.~Wang, and X.~Yang, ``Cross-scene crowd counting via deep
  convolutional neural networks,'' in \emph{Proceedings of the IEEE Conference
  on Computer Vision and Pattern Recognition}, 2015, pp. 833--841.

\bibitem{zhang2016single}
Y.~Zhang, D.~Zhou, S.~Chen, S.~Gao, and Y.~Ma, ``Single-image crowd counting
  via multi-column convolutional neural network,'' in \emph{Proceedings of the
  IEEE Conference on Computer Vision and Pattern Recognition}, 2016, pp.
  589--597.

\bibitem{sindagi2017generating}
V.~A. Sindagi and V.~M. Patel, ``Generating high-quality crowd density maps
  using contextual pyramid cnns,'' in \emph{Proceedings of the IEEE
  International Conference on Computer Vision}, 2017, pp. 1861--1870.

\bibitem{li2018csrnet}
Y.~Li, X.~Zhang, and D.~Chen, ``Csrnet: Dilated convolutional neural networks
  for understanding the highly congested scenes,'' in \emph{Proceedings of the
  IEEE Conference on Computer Vision and Pattern Recognition}, 2018, pp.
  1091--1100.

\bibitem{simonyan2014very}
K.~Simonyan and A.~Zisserman, ``Very deep convolutional networks for
  large-scale image recognition,'' \emph{ArXiv Preprint ArXiv:1409.1556}, 2014.

\bibitem{sajid2020zoomcount}
U.~Sajid, H.~Sajid, H.~Wang, and G.~Wang, ``Zoomcount: A zooming mechanism for
  crowd counting in static images,'' \emph{IEEE Transactions on Circuits and
  Systems for Video Technology}, vol.~30, no.~10, pp. 3499--3512, 2020.

\bibitem{song2021rethinking}
Q.~Song, C.~Wang, Z.~Jiang, Y.~Wang, Y.~Tai, C.~Wang, J.~Li, F.~Huang, and
  Y.~Wu, ``Rethinking counting and localization in crowds: A purely point-based
  framework,'' in \emph{Proceedings of the IEEE/CVF International Conference on
  Computer Vision}, 2021, pp. 3365--3374.

\bibitem{tian2021cctrans}
Y.~Tian, X.~Chu, and H.~Wang, ``Cctrans: Simplifying and improving crowd
  counting with transformer,'' \emph{ArXiv Preprint ArXiv:2109.14483}, 2021.

\bibitem{zhang2022cross}
A.~Zhang, J.~Xu, X.~Luo, X.~Cao, and X.~Zhen, ``Cross-domain attention network
  for unsupervised domain adaptation crowd counting,'' \emph{IEEE Transactions
  on Circuits and Systems for Video Technology}, 2022.

\bibitem{gao2022forget}
J.~Gao, J.~Li, H.~Shan, Y.~Qu, J.~Z. Wang, and J.~Zhang, ``Forget less, count
  better: A domain-incremental self-distillation learning benchmark for
  lifelong crowd counting,'' \emph{arXiv preprint arXiv:2205.03307}, 2022.

\bibitem{nellithimaru2019rols}
A.~K. Nellithimaru and G.~A. Kantor, ``Rols: Robust object-level slam for grape
  counting,'' in \emph{Proceedings of the IEEE/CVF Conference on Computer
  Vision and Pattern Recognition Workshops}, 2019, pp. 1--9.

\bibitem{lins2020method}
E.~A. Lins, J.~P.~M. Rodriguez, S.~I. Scoloski, J.~Pivato, M.~B. Lima, J.~M.~C.
  Fernandes, P.~R.~V. da~Silva~Pereira, D.~Lau, and R.~Rieder, ``A method for
  counting and classifying aphids using computer vision,'' \emph{Computers and
  Electronics in Agriculture}, vol. 169, p. 105200, 2020.

\bibitem{wang2021ssrnet}
D.~Wang, D.~Zhang, G.~Yang, B.~Xu, Y.~Luo, and X.~Yang, ``Ssrnet: In-field
  counting wheat ears using multi-stage convolutional neural network,''
  \emph{IEEE Transactions on Geoscience and Remote Sensing}, vol.~60, pp.
  1--11, 2021.

\bibitem{ranjan2021learning}
V.~Ranjan, U.~Sharma, T.~Nguyen, and M.~Hoai, ``Learning to count everything,''
  in \emph{Proceedings of the IEEE/CVF Conference on Computer Vision and
  Pattern Recognition}, 2021, pp. 3394--3403.

\bibitem{shi2022represent}
M.~Shi, H.~Lu, C.~Feng, C.~Liu, and Z.~Cao, ``Represent, compare, and learn: A
  similarity-aware framework for class-agnostic counting,'' in
  \emph{Proceedings of the IEEE/CVF Conference on Computer Vision and Pattern
  Recognition}, 2022, pp. 9529--9538.

\bibitem{li2017learning}
Z.~Li and D.~Hoiem, ``Learning without forgetting,'' \emph{IEEE Transactions on
  Pattern Analysis and Machine Intelligence}, vol.~40, no.~12, pp. 2935--2947,
  2017.

\bibitem{rebuffi2017icarl}
S.-A. Rebuffi, A.~Kolesnikov, G.~Sperl, and C.~H. Lampert, ``icarl: Incremental
  classifier and representation learning,'' in \emph{Proceedings of the IEEE
  Conference on Computer Vision and Pattern Recognition}, 2017, pp. 2001--2010.

\bibitem{mallya2018packnet}
A.~Mallya and S.~Lazebnik, ``Packnet: Adding multiple tasks to a single network
  by iterative pruning,'' in \emph{Proceedings of the IEEE Conference on
  Computer Vision and Pattern Recognition}, 2018, pp. 7765--7773.

\bibitem{tasar2019incremental}
O.~Tasar, Y.~Tarabalka, and P.~Alliez, ``Incremental learning for semantic
  segmentation of large-scale remote sensing data,'' \emph{IEEE Journal of
  Selected Topics in Applied Earth Observations and Remote Sensing}, vol.~12,
  no.~9, pp. 3524--3537, 2019.

\bibitem{yan2021dynamically}
S.~Yan, J.~Xie, and X.~He, ``Der: Dynamically expandable representation for
  class incremental learning,'' in \emph{Proceedings of the IEEE/CVF Conference
  on Computer Vision and Pattern Recognition}, 2021, pp. 3014--3023.

\bibitem{michieli2019incremental}
U.~Michieli and P.~Zanuttigh, ``Incremental learning techniques for semantic
  segmentation,'' in \emph{Proceedings of the IEEE/CVF International Conference
  on Computer Vision Workshops}, 2019, pp. 1--8.

\bibitem{yang2016wider}
S.~Yang, P.~Luo, C.-C. Loy, and X.~Tang, ``Wider face: A face detection
  benchmark,'' in \emph{Proceedings of the IEEE Conference on Computer Vision
  and Pattern Recognition}, 2016, pp. 5525--5533.

\bibitem{kagglewheat}
``Global wheat detection,''
  \url{https://www.kaggle.com/competitions/global-wheat-detection}.

\bibitem{arteta2016counting}
C.~Arteta, V.~Lempitsky, and A.~Zisserman, ``Counting in the wild,'' in
  \emph{Proceedings of the European Conference on Computer Vision}.\hskip 1em
  plus 0.5em minus 0.4em\relax Springer, 2016, pp. 483--498.

\bibitem{idrees2018composition}
H.~Idrees, M.~Tayyab, K.~Athrey, D.~Zhang, S.~Al-Maadeed, N.~Rajpoot, and
  M.~Shah, ``Composition loss for counting, density map estimation and
  localization in dense crowds,'' in \emph{Proceedings of the European
  Conference on Computer Vision}, 2018, pp. 532--546.

\bibitem{he2016deep}
K.~He, X.~Zhang, S.~Ren, and J.~Sun, ``Deep residual learning for image
  recognition,'' in \emph{Proceedings of the IEEE Conference on Computer Vision
  and Pattern Recognition}, 2016, pp. 770--778.

\bibitem{belouadah2019il2m}
E.~Belouadah and A.~Popescu, ``Il2m: Class incremental learning with dual
  memory,'' in \emph{Proceedings of the IEEE/CVF International Conference on
  Computer Vision}, 2019, pp. 583--592.

\bibitem{castro2018end}
F.~M. Castro, M.~J. Mar{\'\i}n-Jim{\'e}nez, N.~Guil, C.~Schmid, and K.~Alahari,
  ``End-to-end incremental learning,'' in \emph{Proceedings of the European
  Conference on Computer Vision}, 2018, pp. 233--248.

\bibitem{wu2019large}
Y.~Wu, Y.~Chen, L.~Wang, Y.~Ye, Z.~Liu, Y.~Guo, and Y.~Fu, ``Large scale
  incremental learning,'' in \emph{Proceedings of the IEEE/CVF Conference on
  Computer Vision and Pattern Recognition}, 2019, pp. 374--382.

\end{thebibliography}

\vfill

\end{document}